\documentclass[11pt]{article}
\usepackage{amsmath,graphicx,amssymb}
\usepackage{framed}
\usepackage{subfigure}
\usepackage[footnotesize]{caption}
\usepackage{authblk}
\usepackage{fullpage}

\def\dangle{{\theta}}
\def\dangletwo{{\varphi}}
\def\objloc{{\boldsymbol{p}}}
\def\f{{f}}

\def\eone{{\boldsymbol{e}_1}}
\def\etwo{{\boldsymbol{e}_2}}
\def\ethree{{\boldsymbol{e}_3}}

\def\slmpat{\mathbf{\Phi}}

\def\elemx{x}
\def\vecx{\boldsymbol{\elemx}}
\def\specmat{{\boldsymbol{s}}}

\def\elemspas{\psi}
\def\spasvec{\boldsymbol{\elemspas}}
\def\spasis{\boldsymbol{\Psi}}

\def\real{{\mathbb{R}}}

\newcommand{\scp}[2]{\langle #1, #2 \rangle}

\def\sensvec{\bs \Gamma}
\def\sensis{\bs \Gamma}

\DeclareMathOperator*{\argmin}{\arg\min}
\DeclareMathOperator{\diag}{diag}

\def\elemalpha{\alpha}
\def\vecalpha{\boldsymbol{\elemalpha}}
\def\coh{\mu}
\def\senset{\Omega}
\def\measvec{\boldsymbol{y}}
\def\cmplx{\mathbb{C}}
\def\noisevec{\boldsymbol{n}}

\def\noistd{\epsilon}
\def\ssvec{\boldsymbol{m}}
\def\sselem{m}

\def\M{\boldsymbol{M}}
\def\H{\boldsymbol{H}}

\def\gausswdth{\rho}
\def\gaussmat{\mathbf{G}}
\def\gaussvec{{\boldsymbol{g}}}
\def\optrans{\mathcal{T}}
\def\paramtrans{\tau}

\def\allone{\mathbf{1}}

\def\ccdind{k}

\def\sparsity{K}
\def\measdim{M}
\def\sigdim{N}
\def\nccd{N_{\rm C}}
\def\nslm{N}

\newcommand{\loneof}[1]{\|#1\|_1}
\newcommand{\ltwoof}[1]{\|#1\|_2}

\newcommand{\bs}{\boldsymbol}
\newcommand{\bb}{\mathbb}
\newcommand{\ie}{\emph{i.e.}, }
\newcommand{\eg}{\emph{e.g.}, }

\title{  Compressive Schlieren Deflectometry  }

\author[1]{\small P. Sudhakar\thanks{PS is supported by the DETROIT project (WIST3), Walloon Region, Belgium.}}
\author[1]{L Jacques\thanks{LJ is supported by the Belgian FRS-FNRS fund.}}
\author[2]{X. Dubois}
\author[2]{P. Antoine}
\author[2]{L. Joannes}
\affil[1]{ELEN Department, ICTEAM, Universit\'{e} catholique de Louvain, Belgium.}
\affil[2]{Lambda-X, Nivelles, Belgium.}

\date{\small \today}

\begin{document}
\maketitle
\begin{abstract}
Schlieren deflectometry aims at characterizing the deflections undergone by refracted incident light rays at any surface point of a transparent object. For smooth surfaces, each surface location is actually associated with a sparse deflection map (or spectrum). This paper presents a novel method to compressively acquire and reconstruct such spectra. This is achieved by altering the way deflection information is captured in a common Schlieren Deflectometer, \ie the deflection spectra are indirectly observed by the principle of spread spectrum compressed sensing. These observations are realized optically using a 2-D Spatial Light Modulator (SLM) adjusted to the corresponding sensing basis and whose modulations encode the light deviation subsequently recorded by a CCD camera. The efficiency of this approach is demonstrated experimentally on the obsevation of few test objects. Further, using a simple parameterization of the deflection spectra we show that relevant key parameters can be directly computed using the measurements, avoiding full reconstruction. 
\end{abstract}

\section{Introduction}
\label{sec:intro}

Schlieren deflectometry aims at characterizing transparent objects by studying the deflections undergone by refracted incident light rays at any point of their surface point~\cite{Settles:2001}. Compared to other characterization techniques relying on interferometry, deflectometry is also insensitive to vibrations which makes it very attractive for industrial deployment (\eg for quality control).

\begin{figure}[t]
\centering
\raisebox{2mm}{%
\includegraphics[width=.57\columnwidth]{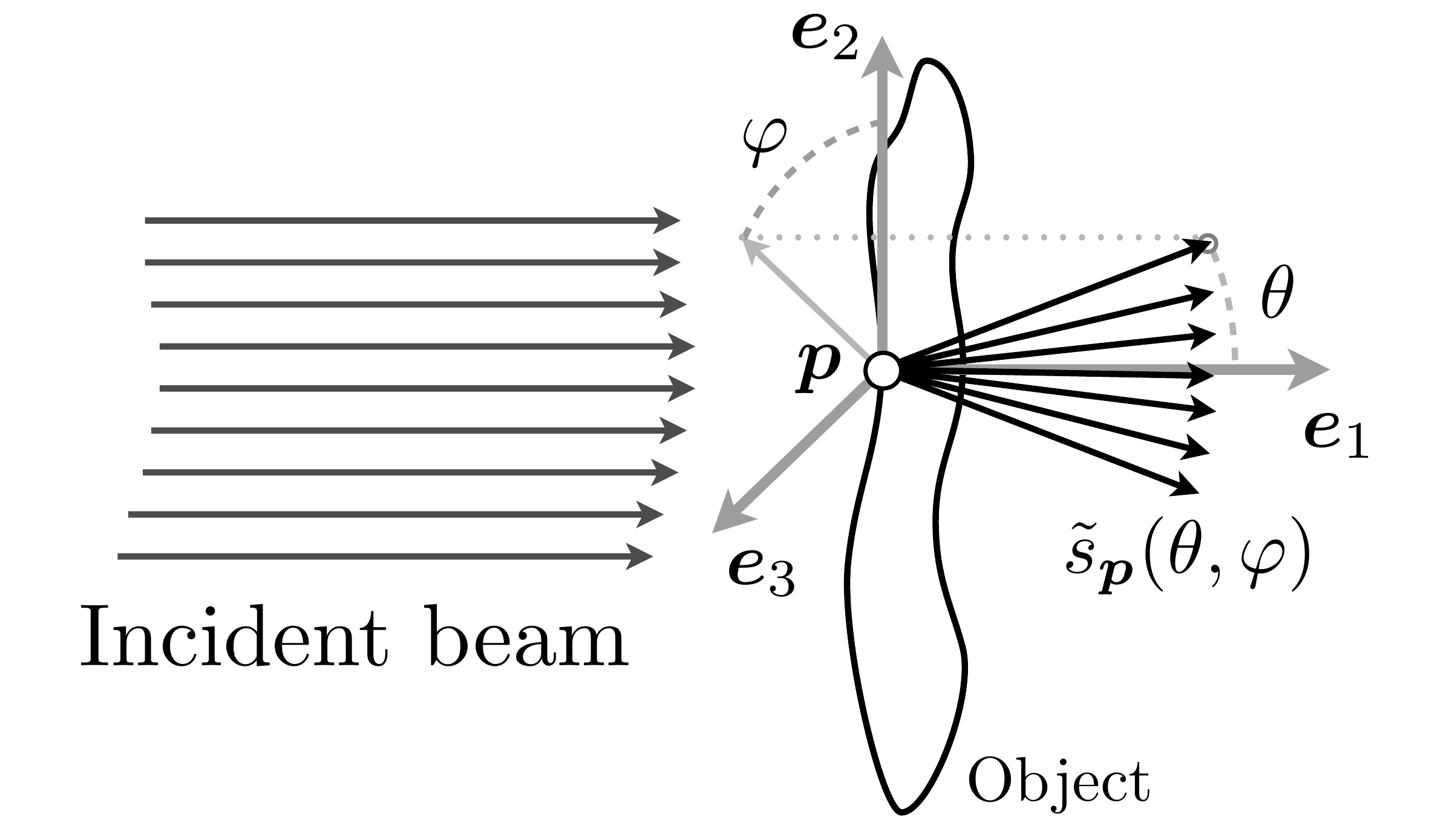}%
}\!\!
{
\includegraphics[width=.4\columnwidth]{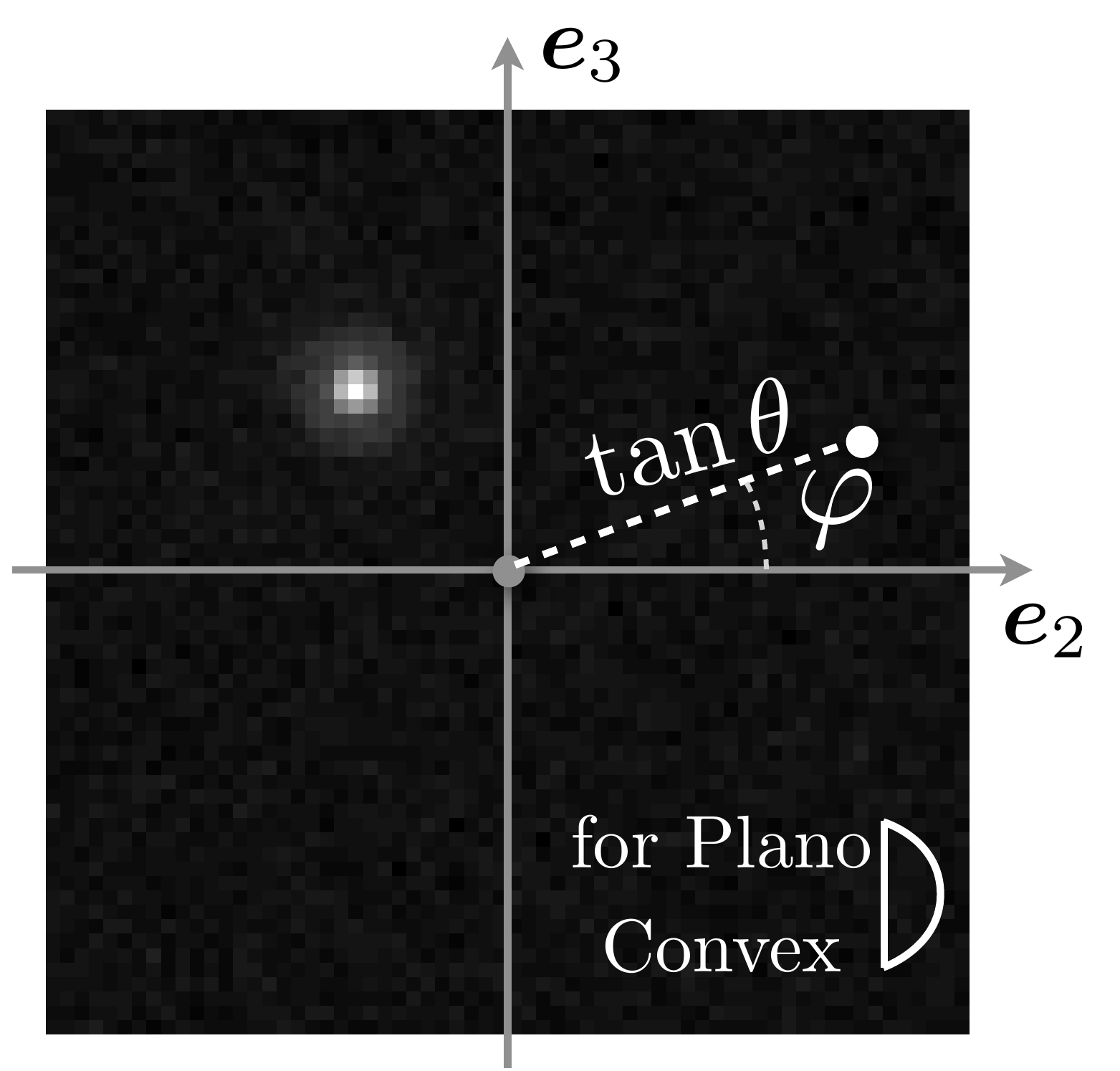}
}
\caption{\label{fig:2dspec-fullsetrecon} Left, illustration of a deflection spectrum. Right, a typical (projected) deflection spectrum $s_{\bs p}$ for a plano convex lens of optical power $25.12$D.}
\end{figure}

Consider a (thin) transparent object shined on one side with a beam of parallel light rays, as shown in Fig.~\ref{fig:2dspec-fullsetrecon}(left). At each surface location $\objloc$, the refracted light is deviated in multiple directions characterized in a local coordinate system $(\eone, \etwo, \ethree)$, with $\bs e_3$ parallel to incident light beam. Using the spherical coordinates $(\dangle, \dangletwo)$ in this system (see Fig.~\ref{fig:2dspec-fullsetrecon}(left)), the resulting \emph{deflection spectrum} $\tilde s_{\bs p}(\dangle, \dangletwo)\in \real_+$ represents the flux of light deviated in the direction $(\dangle, \dangletwo)$. The evolution of deflection spectra along the object surface encodes information about its local curvature. This is why deflectometry is often used for characterizing transparent object surfaces.

In this paper, $\tilde s_{\bs b}$ is conveniently represented by its projection in the $\Pi_{\bs p}=\bs e_2\bs e_3$ plane, \ie according the projected function $s_{\bs p}(r(\theta),\varphi) = \tilde s_{\bs p}(\theta,\varphi)$ with $r(\theta) = \tan\theta$. Moreover, the object surface is assumed sufficiently smooth for being parametrized by a projection of $\bs p$ in the same plane (on a arbitrary fixed origin), so that $\bs p$ is basically a 2-D vector. 

For most objects (\eg with smooth surfaces) deflections at any location $\objloc$ occur in a limited range of angles. The deflection spectra therefore tend to be naturally \emph{sparse} in plane $\Pi_{\bs p}$ or in some appropriate basis of this domain (\eg wavelets). Fig.~\ref{fig:2dspec-fullsetrecon}(right) shows an example of a discretized deflection spectrum $s_{\bs p}$ for one location of a plano convex lens obtained using the setup described in Sec.~\ref{sec:setup}. The white spot in the image signifies that deflections occur at only a few angles (as governed by classical optics) and deflections elsewhere are negligible. 

Measuring a deflection spectrum $s_{\bs p}$ for every $\objloc$ is not straightforward. The optical setup described in Sec.~\ref{sec:setup} measures them indirectly
by optical comparison with a certain number of programmable modulation patterns. To maximize the amount of information collected in this process, we exploit the sparse nature of spectra in each plane $\Pi_{\bs p}$ and use the framework of \emph{spread spectrum}\footnote{``Spread Spectrum'' is not related to the studied deflection ``spectrum'' but it refers to the signal frequency spectrum.} compressive sensing~\cite{Puy:2011p1751}, described in Sec.~\ref{sec:sscs}. From a fine calibration of the system relative to its intrinsic noise, our approach is then experimentally validated using deflectometric measurements and the numerical spectrum reconstruction results are presented in Sec.~\ref{sec:reconresults}. Further, from the possibility of performing signal processing in the compressed domain~\cite{Davenport:2007p2548, Davenport:2010p589}, we also show in Sec.~\ref{sec:specchar} that for localized deflection spectra summarized by a few optical parameters (\eg location and width of the main peak) the compressed optical observations themselves can be processed. This is aimed at extracting the relevant information at a smaller number of measurements than those required for a full spectrum reconstruction.

\begin{figure}[t]
\centering
\includegraphics[width=\columnwidth]{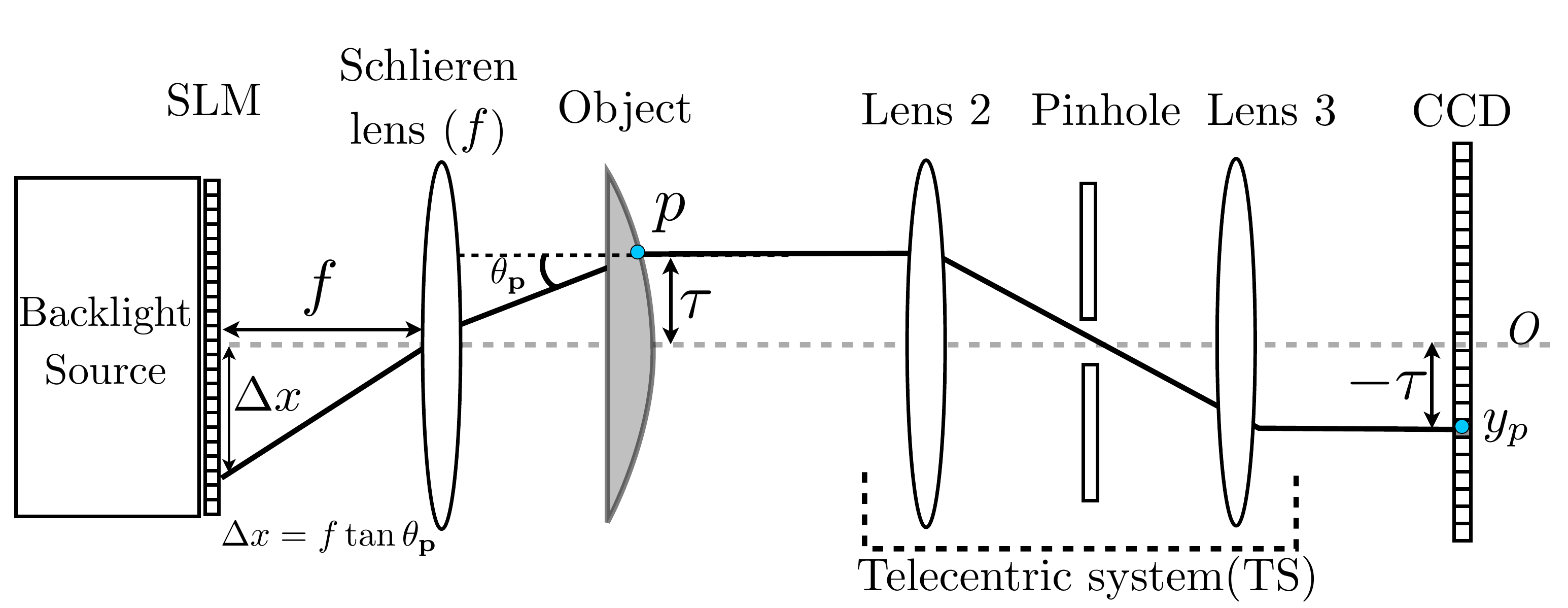}
\caption{\label{fig:pss} A 2-D schematic of Schlieren deflectometer.  }
\end{figure}

\section{Optical setup and Notations}
\label{sec:setup}

Deflection spectra can be measured by the Schlieren deflectometer represented in Fig.~\ref{fig:pss}. Its key components are \emph{(i)} a Spatial Light Modulator (SLM), \emph{(ii)} the Schlieren lens with focal length $\f$, \emph{(iii)} the Telecentric System (TS) and \emph{(iv)} the Charged Coupled Device (CCD) camera collecting the light. 

The object to be analyzed is placed in between the Schlieren lens and the telecentric system. It is shined on its left by a light source and, thanks to the telecentric system, only the parallel light rays emerging out from the object are collected by the CCD. Moreover, up to a flipping around the optical axis, each location $\objloc$ on the object, at a distance $\tau$ from the optical axis $O$ (dashed line), is probed by a corresponding CCD pixel also at a distance of $\tau$ from $O$. Each location $\bs p$ is thus in one-to-one correspondence with a CCD pixel and we will sometimes consider $\bs p$ as CCD pixel location. 

From classical optics, a light ray that is incident on location $\objloc$ at an angle $\dangle_\objloc$ originates from the light source at a distance of $\Delta x = \f\tan\dangle_\objloc$ from the optical axis. Likewise, the light rays originating at different locations on the source have different incident angles at $\objloc$. Since we can always virtually invert the light propagation in the system, everything works as if the object was shined on its right by a beam of parallel light rays. 
Therefore, up to a global scaling by $f$, the SLM plane is actually the local plane $\Pi_{\bs p}$  of the deflection spectrum occurring at $\bs p$. Modulating the SLM corresponds to modulating $s_{\bs p}$, while the light collected in CCD pixel $\bs p$ is just an inner product of $s_{\bs p}$ with the modulation. 

If we generate $M$ such modulations $\phi_{i}\in\bb R^\nslm$ with $1\leq i \leq M$ in the SLM of $N$ pixels, considering the discrete nature of the CCD camera (having $\nccd$ pixels), the discretized deflection spectra are observed through 
\begin{equation}
  \label{eq:schlieren-forward-model}
  \bs y_k = \bs \Phi \bs s_k + \bs n,\quad 1\leq k \leq \nccd,
\end{equation}
where $\bs \Phi^T = (\bs \phi_1, \cdots, \bs \phi_M) \in \bb R^{N \times M}$ is the sensing matrix, $k$ is a CCD pixel index, $\bs s_k\in\bb R^N$ is the discretized spectrum at the $k^{\rm th}$ pixel/object location, and $\bs n$ models the measurement noise (assumed Gaussian). Notice that the SLM and the CCD 2-D grids are represented as 1-D spaces in order to ease the notation, so that $\bs \Phi$ is then a sensing 2-D matrix acting on 1-D vectors. 

The quest now is to optimize the design of $\bs \Phi$ in order to
maximize the spectrum information captured in each $\bs y_k$, \ie
using Compressed Sensing theory.

\section{Prior work}

A well known example of inner products based imaging system is the Rice University's \emph{single pixel camera}~\cite{Duarte:2008}. This camera uses a single photosensor to capture the inner products between the scene to be imaged and random binary modulation patterns, realised using a micro-mirror array.  In our deflectometric system, each CCD pixel behaves like a single pixel camera, which poses a great computational challenge for reconstruction. We can notice also that a binary SLM modulation for compressive imaging is also advocated in \cite{romberg2009compressive} for performing random convolutions of images. However, this method modulates the Fourier transform of the signal of interest and not its spatial domain representation.  

The previous use of Schlieren Deflectometer is in the Phase Shifting Schlieren (PSS) system. PSS quantitively measures the deviation angles from the deflectometric measurements, by assuming single deflection for each object location. It overcomes several calibration issues of the traditional deflectometry by using multiline phase shifted spatial filters in the SLM~\cite{Joannes:2003p1444}. In this case, the CCD pixel values $\measvec_\ccdind$ directly encode the deviation angle of the corresponding object location $\ccdind$. By making several measurements with phase shifted sinusoidal modulations as $\bs \Phi$, the deviation angle is then numerically decoded from the $\measvec_\ccdind$, using $n$-step algorithms~\cite{Hariharan:87}. PSS has been successfully used in tomographic applications such as refractive index map reconstruction~\cite{Beghuin:2010p2208, AGonzalez:2012}. However, the phase shifting method is unable to recover the full deflection spectra or to estimate several main deflection angles per point. Moreover, the use of non-binary modulation patterns brings in the problem of non-linearity in the SLM response. This is one of the reasons why we advocate using binary modulation patterns as described in Sec.~\ref{sec:reconresults}.

\section{Spread Spectrum Compressive Sensing}
\label{sec:sscs}

Compressive Sensing (CS) shows that any signal $\vecx = \spasis\vecalpha\in\cmplx^\sigdim$, having a {\em sparse} representation in an orthonormal {\em sparsity basis} $\spasis\in\cmplx^{\sigdim\times\sigdim}$, \ie $\|\vecalpha\|_0:=\#\{j : \elemalpha_j\neq 0\}\leq \sparsity \ll N$, can be tractably recovered from a few corrupted linear measurements of the form
\begin{equation}
\label{eqn:csmeasurement}
\measvec = \sensis_\senset^\ast \spasis \vecalpha\; + \;\noisevec,
\end{equation}
where $^*$ denotes the conjugate transpose, $\sensis\in\cmplx^{\sigdim\times\sigdim}$ is an orthonormal \emph{sensing basis}, $\sensis_\senset$ is the submatrix formed by restricting the columns of $\sensis$ to those belonging to $\senset\subset [N] :=\{1,\,\cdots, N\}$ and $\noisevec$ is a Gaussian noise vector. 

In particular, if $\Omega$ has size $M$ and is drawn uniformly at random in $[N]$, and if $\sensis$ and $\spasis$ are incoherent, meaning that 
the \emph{coherence} $\coh :=\sqrt{\sigdim} \max_{1\leq i,j\leq \sigdim} |\langle \sensvec_j, \spasvec_i\rangle|$ is very close to 1, then, 
\begin{equation*}
\measdim = O\big(\coh^2\sparsity\log^4(\sigdim)\big)
\end{equation*}
measurements are enough to recover a good estimate of $\bs x$~\cite{Rauhut:2010p2556}. This reconstruction is performed by solving the following convex optimization
\begin{equation}
\label{eqn:bpdn}
\widehat{\bs\alpha} := \argmin_{\widetilde\vecalpha\in\cmplx^\sigdim}\;\loneof{\widetilde\vecalpha}\;\text{subject to}\; \ltwoof{\measvec - \slmpat \widetilde\vecalpha}\leq\noistd,
\end{equation}
where $\bs \Phi = \bs \Gamma^*_{\Omega}\bs\Psi$, and $\epsilon$ is a bound on $\|\bs n\|_2\leq \epsilon$. Under the previous conditions, the theory guantees that~\cite{candes2007sparsity}
\begin{equation}
\label{eqn:bpdnreconerr}
\ltwoof{\vecalpha - \widehat{\vecalpha}} = O\big(\tfrac{\loneof{\vecalpha - \vecalpha_\sparsity}}{\sqrt{\sparsity}}\;+\;\noistd\big)
\end{equation}
holds with a probability at least $1-\sigdim^{-\gamma\log^3(\sigdim)}$, where $\vecalpha_\sparsity$ is the best $\sparsity$-term approximation of the vector $\vecalpha$.

As shown above, the number of the measurements scales quadratically as $\coh$ and hence it is desirable to have the two bases as incoherent as possible to make $\coh$ close to $1$. In order to circumvent this problem, \cite{Puy:2011p1751} cleverly makes the sensing and sparsity bases incoherent by a simple pre-modulation, or \emph{spread spectrum}, of the data vector $\vecx$. This technique is optimal for \emph{universal} sensing basis, \ie when all the entries of $\bs \Phi$ have the same complex amplitude, which is the case for Fourier and Hadamard bases. 

Mathematically, with random modulation, the sensing matrix becomes $\bs \Phi = \sensis_\senset^\ast\M \spasis$, where $\M=\diag(\bs m)$ is a diagonal matrix whose diagonal is a random vector $\ssvec$ such that $|\sselem_i|=1$, \eg a Steinhaus or Rademacher sequence. In this case, we need
\begin{equation*}
\measdim\geq C_\rho\,\sparsity\log^5(\sigdim) 
\end{equation*}  
measurements in order to recover a solution $\vecalpha^\star$ of~\eqref{eqn:bpdn} satisfying~\eqref{eqn:bpdnreconerr} with a probability at least
$1-O(\sigdim^{-\rho})$, for some $0<\rho<\log^3(\sigdim)$. Noticeably, the coherence has disappeared from the condition implying that with spread spectrum and universal sensing basis, the recovery guarantee is {\em universal}, irrespective of the sparsity basis.  We see in next section how to exploit the spread spectrum method in our optical setup.

\section{Deflection Spectrum Reconstruction}
\label{sec:reconresults}

We propose to combine the Schlieren deflectometer of
Sec.~\ref{sec:setup} with the spread spectrum CS principles described
above.\\[-3mm] 

\noindent\textbf{Optical Sensing:} Since it is essential to have real valued sensing basis and
spread spectrum vector, as allowed by \cite{Puy:2011p1751}, we use the Hadamard (universal) basis $\H$ for sensing, with $\bs H \bs H^T = \bs H^T \bs H = \bs{\rm Id}$ and $H_{ii'}\in\{\pm \frac{1}{rt N}\}$, combined with a random Rademacher vector $\ssvec$, \ie $m_i=\pm 1$ independently with equal probability. 

Since the SLM accepts only positive values, once the random set $\Omega \subset [N]$ has been drawn, the corresponding sensing matrix $\bs \Phi = \H^T_{\Omega} \M$ must be 
suitably \emph{biased} and scaled as
\begin{equation}
\label{eqn:snesmatscale}
\overline{\bs \Phi} = \tfrac{1}{2}(\bs \Phi + \allone_\sigdim\allone_\sigdim^T) \in \{0,1\}^{M\times N},
\end{equation}
where $\allone_\sigdim \in \{1\}^N$ is the vector of ones. 
The choice of Hadamard makes the biased sensing binary. This has
interesting advantages in the optical setup since it avoids any
non-linear response of the SLM for non-binary values. 

Moreover, thanks to an extra measurement of each $\specmat_\ccdind$, \ie the value $\overline z_k =
\scp{\bs 1_N}{\bs s_k}$ obtained by an entirely transparent SLM,  any
measurement vector $\overline{\bs y}_k = \overline{\bs \Phi} \bs s_k$
can be debiased by  

\begin{equation}
\label{eqn:measvecdb}
\measvec_\ccdind := \slmpat \specmat_\ccdind = 2\overline{\measvec}_\ccdind - \overline{z}_k\,\allone_\measdim. 
\end{equation} 
Up to a correct evaluation of the noise, which corrupts $\overline{\bs y}_k$, ${\bs z}_k$ and the debiasing process \eqref{eqn:measvecdb}, we can therefore reconstruct $\specmat_\ccdind$ from the unbiased $\measvec_\ccdind$.\\[-3mm]

\noindent
\textbf{Noise estimation:} If there is no test object, then by classical optics the measured deflection spectrum is constant in all CCD pixels and corresponds to a simple disk centered on the origin of the spectrum domain. We denote it as $\specmat^{\rm no}$. The disk diameter is proportional to the pinhole diameter of the system (see Fig.~\ref{fig:pss}). This prior information aids us in calibrating the system and in estimating the noise level on the measurements. 

From actual measurements in the absence of test object, we obtain, on an arbitrary CCD pixel, $\measvec^{\rm no} = \slmpat(\specmat^{\rm no} + \noisevec_{\specmat}) + \noisevec_{\measvec}$, where $\noisevec_{\specmat}$ and $\noisevec_{\measvec}$ are the unknown signal and observation noises. After a small calibration of the SLM origin, and up to a small optimization of the disk height in $\specmat^{\rm no}$, we can therefore compute a bound on the noise power as $\epsilon = \|\slmpat \noisevec_{\specmat} + \noisevec_{\measvec}\| =  \|\measvec^{\rm no} - \slmpat\specmat^{\rm no}\|$.   We can either obtain this value for every $M$ or estimate it for $M=N$ only and scale the result as $\epsilon(N)=\sqrt{\measdim+2\sqrt{\measdim}}\ \epsilon(M)/\sqrt{N}$ for $\measdim<\sigdim$. \\[-3mm]

\noindent\textbf{Reconstruction Method:} For the reconstruction, we select the Daubechies 9/7 wavelet basis as our sparsity basis~\cite{Mallat:2008:WTS:1525499}. This offers a sparser representation of the deflection spectra than the canonical (Dirac) basis. 
To reconstruct the deflection spectrum at any location $\ccdind$, an estimate of the sparse wavelet coefficients $\widehat{\vecalpha}_\ccdind$ is obtained by solving~\eqref{eqn:bpdn} with the $\epsilon$ estimated above. The spectrum is then estimated by $\hat{\specmat}_\ccdind = \spasis^\ast\widehat{\vecalpha}_\ccdind$. To solve~\eqref{eqn:bpdn}, we used the Chambolle-Pock algorithm, a first order primal-dual method for solving convex optimization problems using proximal operators~\cite{Chambolle:2011p1809}.

For evaluating compressive reconstruction performance, \eqref{eqn:bpdn} was solved with $\measdim = \sigdim$ measurements to obtain the reference reconstruction $\tilde{\specmat}_\ccdind$. Reconstructions for $M<N$ were compared with $\tilde{\specmat}_\ccdind$ using the (output) Signal-to-Noise Ratio $\text{oSNR} := 20\log_{10}(\ltwoof{\tilde{\specmat}_\ccdind}/\ltwoof{\tilde{\specmat}_\ccdind - \hat{\specmat}_\ccdind})$. \\[-3mm]

\noindent\textbf{Experimental Results:} For experimental evaluation, we considered two plano convex lenses of optical powers $10.03$D and $60$D, and restricted the size of spectrum to $64\times 64$ centered around the SLM origin, resulting in $\sigdim=4096$. For 5 CCD locations, 10 independent reconstruction trials were performed for several values of $\measdim$, by randomly drawing a new $\senset\subset [N]$ every time. 

\begin{figure}[t]
\centering
\includegraphics[scale=0.5]{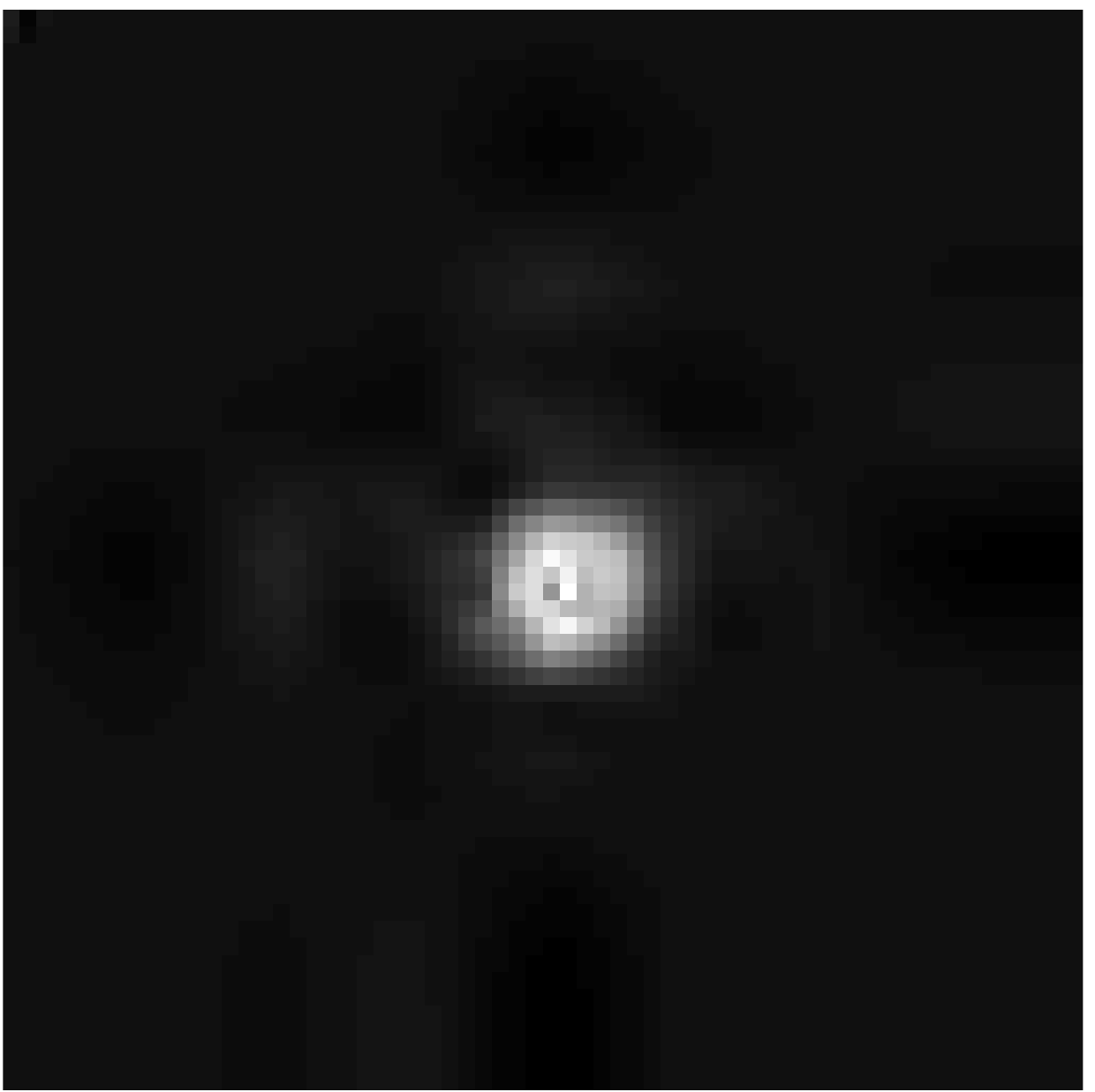}
\includegraphics[scale=0.5]{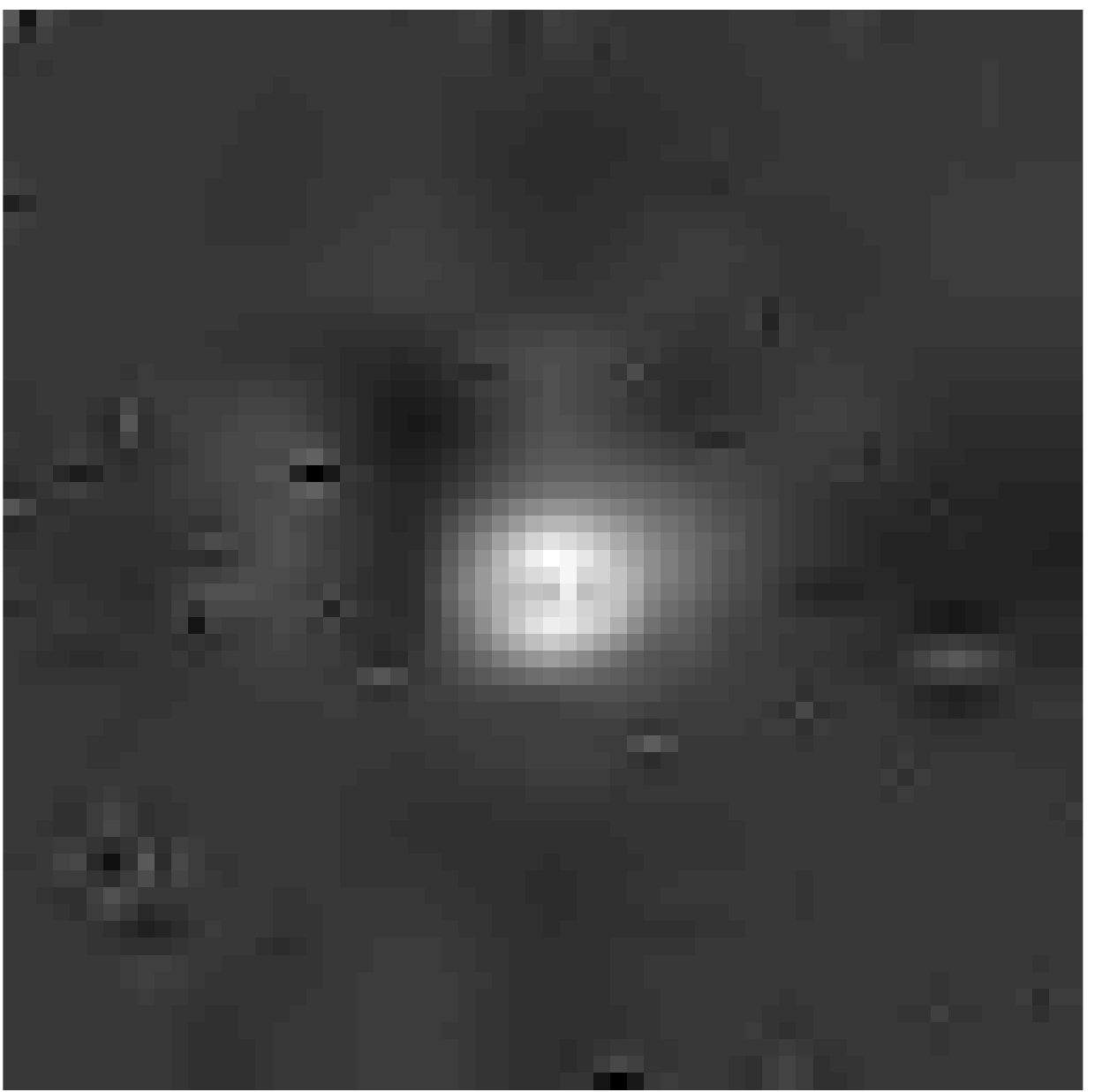}  
\caption{Left, an example of reconstruction using full set of measurements and right, only 10\% of measurements.  }
\label{fig:reconexamples}
\end{figure}

\begin{figure}
\centering
\includegraphics[width=0.8\columnwidth]{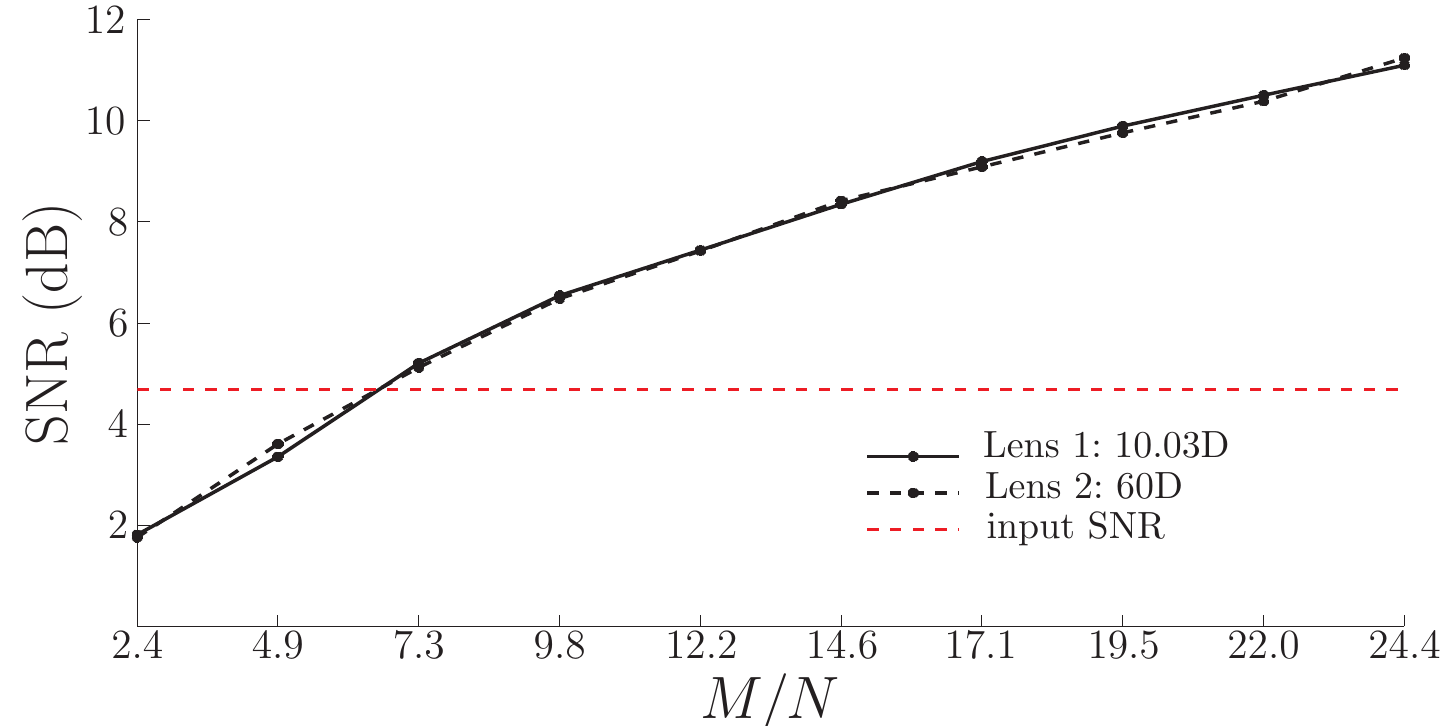}
\caption{Average reconstruction oSNR (in dB) as a function of $M/N$. }
\label{fig:reconResults}
\end{figure}

Fig.~\ref{fig:reconexamples}(left) shows a deflection spectrum reconstructed using the full set of measurements for the lens with $60D$ optical power and Fig.~\ref{fig:reconexamples}(right) shows the reconstruction with $400$ measurements ($M/N\approx10\%$). Note that the spectrum is well localized, corroborating our initial observation.
Fig.~\ref{fig:reconResults} shows the plot of oSNR versus the number of measurements $\measdim/\sigdim$ (in \%), averaged over the trials and locations. The performance curves for both the lenses almost overlap, and the oSNR improves as $\measdim/\sigdim$ increases. Though the absolute values of oSNR seem low, its significance has be understood in the light of the input SNR, which is approximately computed as $\text{iSNR}\ := 20\log_{10}(\ltwoof{\slmpat\specmat^{\rm no}}/\ltwoof{\measvec^{\rm no} - \slmpat\specmat^{\rm no}}) \simeq 4.34$ dB. The horizontal dotted line on the plot indicates the iSNR for our experiments, and it is clear that the reconstruction procedure improves the oSNR, beyond the iSNR, thereby demonstrating the ability of CS reconstruction of deflection spectra in low input SNR regime.  

\section{\hspace{-5pt} Deflection spectrum characterization}
\label{sec:specchar}

Suppose we are interested in summarizing the deflection spectra by a few parameters that characterize them at each location. For example, we could summarize a spectrum by the location of the bright feature it contains. To address this, we propose the following robust method to localize the feature. 

Assuming that the feature is circular and its radius $\gausswdth$ can be known a priori, the feature can be localized using a matched filter. To this end, we employ a template $\gaussmat_\gausswdth$, which contains a two-dimensional Gaussian of width $\gausswdth$ at its center. To localize the feature in a spectrum $\tilde{\specmat}_\ccdind$ we simply translate $\gaussmat_\gausswdth$ over the support of the spectrum and find the best translation $\hat{\paramtrans}\in\real^2$ that maximizes the correlation between the spectrum and the translated template $\optrans_{{\paramtrans}}\gaussmat_\gausswdth$. By letting $\gaussvec^\gausswdth_\paramtrans$ to denote the vectorized form of $\optrans_{{\paramtrans}}\gaussmat_\gausswdth$, we solve

\begin{equation}
\label{eqn:centroid}
\hat{\paramtrans}_\ccdind = \arg\max_{\paramtrans}\,|\langle\tilde{\specmat}_\ccdind,\gaussvec^\gausswdth_\paramtrans\rangle|.
\end{equation}

With an abused terminology we designate the translation parameters $\hat{\paramtrans}$ as the \emph{centroids} of the spectral features. These centroids also provide local information about the mean deflection angles at various locations.

In order to characterize an object using the centroids defined in~\eqref{eqn:centroid}, we have to reconstruct the spectra at all $\nccd$ locations individually and then compute the centroids, which requires huge computational time. It essential therefore to explore the possibility of computing the centroids right from the measurements $\measvec_\ccdind$, without even reconstructing the spectra. 

Davenport et al. have proposed the idea of performing common signal processing tasks such as detection and classification using the compressive samples obtained using random measurements~\cite{Davenport:2007p2548, Davenport:2010p589}. In the same spirit, we compute the centroids directly from the measurements $\measvec_\ccdind$ defined in~\eqref{eqn:measvecdb}. To accomplish this {\em compressive centroid estimation}, we carry the definition of the centroid of the spectrum~\eqref{eqn:centroid} into the measurement domain and solve

\begin{equation}
\label{eqn:compcentroid}
\widetilde{\paramtrans}_\ccdind = \arg\max_{\paramtrans}\,|\langle\slmpat^T\measvec_\ccdind, \gaussvec^\gausswdth_\paramtrans\rangle|.
\end{equation} 

Notice that solving~\eqref{eqn:compcentroid} is the same as solving~\eqref{eqn:centroid}, but instead of using $\tilde{\specmat}_\ccdind$, obtained by solving~\eqref{eqn:bpdn}, we simply compute $\slmpat^T\measvec_\ccdind$ and use it for centroid estimation. 

For experimental evaluation, we retained the configuration described in Sec.~\ref{sec:reconresults}. For each of the 5 CCD locations $\ccdind$, the centroid $\widetilde{\paramtrans}_\ccdind$ was computed by solving~\eqref{eqn:compcentroid}. A ``ground truth" centroid $\hat{\paramtrans}_\ccdind$ was also found by solving~\eqref{eqn:centroid}, with $\specmat_\ccdind$ reconstructed (solving~\eqref{eqn:bpdn}) using full $\measdim = 4096$ measurements.

\begin{figure}
\centering
\includegraphics[width=0.8\columnwidth]{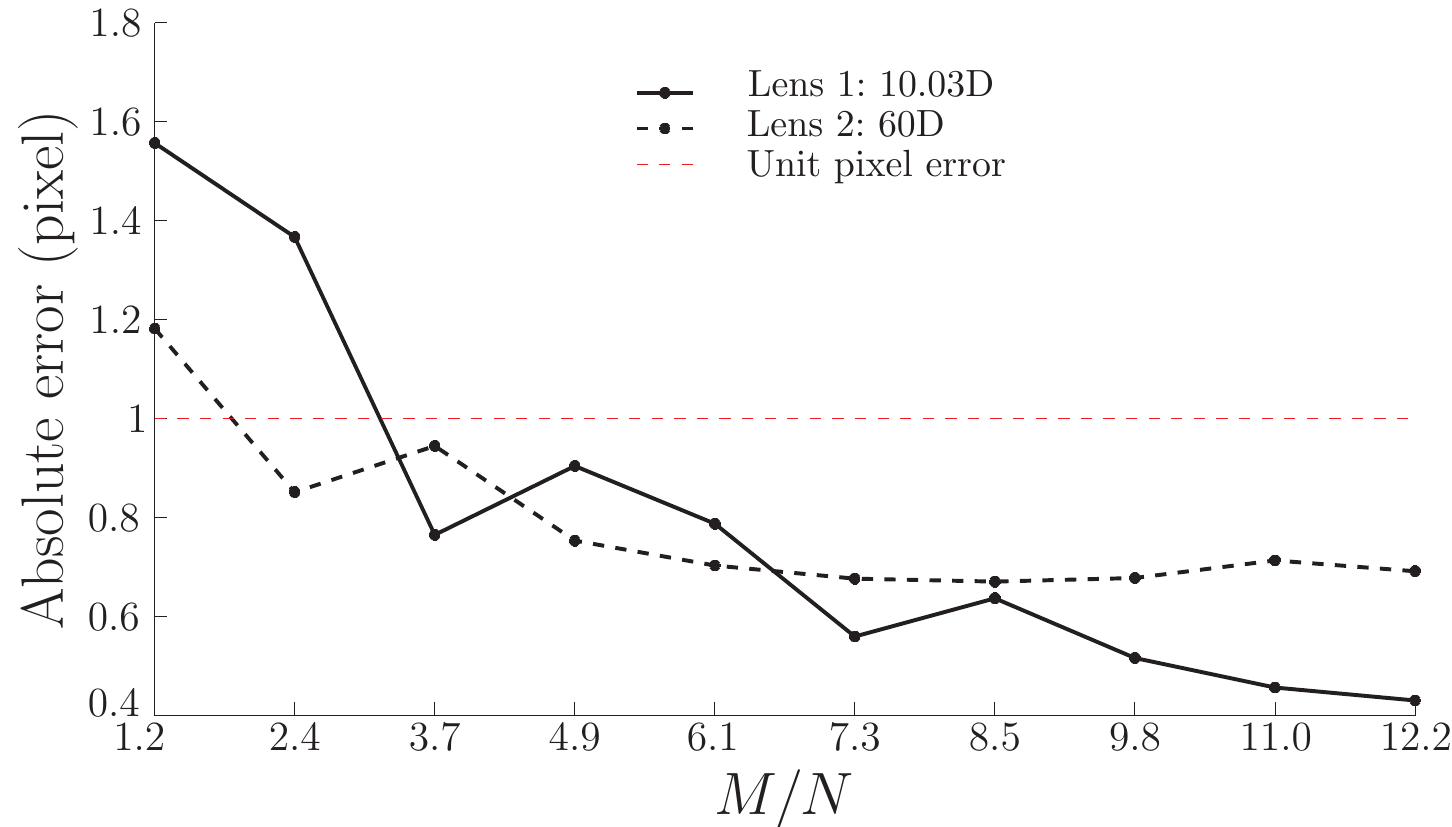}
\caption{Average error in centroid estimation (in pixels) using compressive measurements, as a function of number of measurements, in \%.}
\label{fig:compcentroid}
\end{figure}

Fig.~\ref{fig:compcentroid} shows the centroid computation error $\ltwoof{\hat{\paramtrans} - \tilde{\paramtrans}}$ as a function of the number of measurements $\measdim/\sigdim$. Each data point is an average over 50 independent trials for each value of $\measdim$ and over all the locations. The horizontal dotted line indicates a unit pixel error and it can be seen that the compressive centroid estimation achieves sub-pixel accuracy, even with the number of measurements as low as $2.4\%$ ($50$ measurements) for the $60$D lens, and about $3.7\%$ for the $10.03$D lens. This demonstrates the ability of the measurements to capture sufficient information about the feature of the spectra. 

\section{Conclusions and perspectives}

This paper presents a novel approach to perform schlieren deflectometry using compressive sensing principles. In contrast to the existing system, the proposed approach enables the reconstruction of deflection spectra, instead of a single deflection angle at each location. Although it involves computationally intensive procedure, spectra reconstruction leads to better characterization of the studied object. The empirical results also demonstrate the capability of our method to perform meaningful inferences about the deflection spectra using only the compressive measurements, without involving any reconstruction.  

Several challenges remain to be addressed in the future work. It is of foremost importance to fully understand the noise model to tune the reconstruction algorithm. Also, incorporating further priors such as non-negativity helps the reconstruction. We also intend to develop approaches to simultaneously reconstruct spectra, corresponding to neighbouring CCD pixels, by exploiting the correlation amongst them.

\end{document}